%% file: main.tex
\pdfoutput=1

\RequirePackage{snapshot}
\documentclass[11pt]{article}

\usepackage[]{acl2023}

\usepackage{times}
\usepackage{latexsym}

\usepackage[T1]{fontenc}

\usepackage[utf8]{inputenc}

\usepackage{microtype}

\usepackage[]{todonotes}

\newcommand{\draftonly}[1]{#1}
\renewcommand{\draftonly}[1]{}

\input{latex/packages}

\input{latex/commands_math}

\input{latex/commands_custom}

\input{latex/comments}

\title{Toward Interactive Dictation}

\author{\
Belinda Li$^\text{\ding{171}}$\thanks{\ \ \  Work performed during a research internship at Microsoft Semantic Machines.}%
\ \ \ \ \ 
Jason Eisner$^\Diamond$%
\ \ \ \ \ 
\textbf{Adam Pauls$^\Diamond$}%
\ \ \ \ \ 
\textbf{Sam Thomson$^\Diamond$}%
\\$^\text{\ding{171}}$MIT CSAIL \hspace{1cm} $^\Diamond$Microsoft Semantic Machines \\ 
$^\text{\ding{171}}$\texttt{bzl@mit.edu} \\
$^\Diamond%
$\texttt{\{jason.eisner,adam.pauls,samuel.thomson\}@microsoft.com} \\
}

\begin{document}

\maketitle
\begin{abstract}
\input{sections/0-abstract}

\end{abstract}

\input{sections/1-intro}

\input{sections/2-background}

\input{sections/3-task}

\input{sections/4-dataset}

\input{sections/5-models}
\input{sections/6-results}

\input{sections/7-conclusion}

\section{Limitations}
\label{sec:limitations}
\ourdataset is a pilot dataset.
In particular, its test set can support \segment-level metrics, but is not large enough to support reliable dialogue-level evaluation metrics.
Due to resource constraints, we also do not report inter-annotator agreement measurements.
While we made effort to make our interface low-friction, the demonstration setting still differs from the test-time scenario it is meant to emulate, and such a mismatch may also result in undesired data biases.
Because our dialogues were collected before having a trained interpretation model, trajectories always follow gold interpretations.
Because of this,
the main sources of errors are ASR misdetections or user speech errors.
In particular,~\ourdataset contains data on: 1. misdetections and speech errors in transcription, and how to fix them through commands, 2. misdetections and speech errors in edits, and what intent they correspond to. We leave to future work the task of addressing semantic errors and ambiguities which result from incorrect interpretation of user intent.
Some of these limitations can be addressed by incorporating trained models into the demonstration interface, which will allow faster demonstration, and capture trajectories that include actual system (non-gold) interpretations.

Though the trained system runs, we have not done user studies with it because it is not production-ready.  The T5-base models are efficient enough,
but the prompted GPT3 model is too slow for a responsive interactive experience.
Neither model is accurate enough at interpretation.  We welcome more research on this task!

When a human dictates to another human, interleaved corrections and commands are often marked prosodically (by pitch melody, intensity, and timing).  Our current system examines only the textual ASR output; we have given no account of how to incorporate prosody, a problem that we leave to future work. We also haven't considered how to make use of speech lattices or $n$-best lists, but they could be very useful if the user is correcting our mistranscription---both to figure out what text the user is referring to, and to fix it.

\section{Impact Statement}
This work makes progress toward increasing accessibility for those who cannot use typing inputs.
The nature of the data makes it highly unlikely that artifacts produced by this work could be used 
(intentionally or unintentionally) to quickly generate factually incorrect, hateful, or otherwise malignant text.

The fact that all speakers in our dataset were native speakers of American English
could contribute to exacerbating the already present disparity in usability for English vs. non-English speakers.
Future work should look to expand the diversity of languages, dialects, and accents covered.

\bibliography{main}
\bibliographystyle{acl_natbib}

\clearpage
\newpage
\appendix

\input{sections/999-appendix}

\end{document}

%% file: latex/packages.tex
\usepackage{booktabs}
\usepackage{multirow}
\usepackage{graphicx}

\usepackage{pifont}
\usepackage{amsmath}
\usepackage{amssymb}
\usepackage{ifthen}
\usepackage{dsfont}

\usepackage{listings}
\usepackage{xcolor}
\usepackage{ltablex}
\usepackage{longtable}
\usepackage{multicol}
\usepackage{floatrow}

\usepackage{tikz}
\usetikzlibrary{bayesnet}
\usepackage{subfloat}
\usepackage{subfig}
\usepackage[linguistics]{forest}

\usepackage[ruled,vlined]{algorithm2e}

\usepackage{caption}

\usepackage{cancel}

\usepackage{xspace}
\usepackage[inline,shortlabels]{enumitem}
\usepackage{comment}
\usepackage{color,soul}

\usepackage[noabbrev,capitalize]{cleveref}

\crefname{page}{page}{pages}
\crefname{footnote}{footnote}{footnotes}
\crefname{equation}{equation}{equations}
\crefname{line}{line}{lines}
\crefrangeformat{line}{lines #3#1#4--#5#2#6}
\crefname{lstlsting}{Listing}{Listings}   
\crefname{section}{\S}{\S\S}
\Crefname{section}{\S}{\S\S}
\crefformat{section}{#2\S#1#3}
\Crefformat{section}{#2\S#1#3}
\crefrangeformat{section}{\S\S#3#1#4--#5#2#6}
\Crefrangeformat{section}{\S\S#3#1#4--#5#2#6}
\crefmultiformat{section}{\S#2#1#3}{ and~\S#2#1#3}{, \S#2#1#3}{ and~\S#2#1#3}
\Crefmultiformat{section}{\S#2#1#3}{ and~\S#2#1#3}{, \S#2#1#3}{ and~\S#2#1#3}
\crefrangemultiformat{section}{\S\S#3#1#4--#5#2#6}{ and~\S\S#3#1#4--#5#2#6}{, \S\S#3#1#4--#5#2#6}{ and~\S\S#3#1#4--#5#2#6}
\Crefrangemultiformat{section}{\S\S#3#1#4--#5#2#6}{ and~\S\S#3#1#4--#5#2#6}{, \S\S#3#1#4--#5#2#6}{ and~\S\S#3#1#4--#5#2#6}

%% file: latex/commands_math.tex
\newcommand{\prob}[2][]{p\ifthenelse{\not\equal{}{#1}}{_{#1}}{}(#2)}
\newcommand{\expect}[2][]{\text{\bf E}\ifthenelse{\not\equal{}{#1}}{_{#1}}{}\!\left[#2\right]}
\newcommand{\var}[2][]{\text{\bf Var}\ifthenelse{\not\equal{}{#1}}{_{#1}}{}\!\left[#2\right]}

%% file: latex/commands_custom.tex
\newcommand{\defn}[1]{{\bf{#1}}}
\newcommand{\ie}{\emph{i.e.}, }
\newcommand{\eg}{\emph{e.g.}, }

\newcommand{\ourdataset}{TERTiUS\xspace}
\newcommand{\taskname}{interactive dictation}

\newcommand{\allUttsConcated}{\mathcal{U}}

\newcommand{\enronTask}{Replicate doc}
\newcommand{\emailExpandTask}{Elaborate doc}
\newcommand{\replicateOpTask}{Replicate \op}

\newcommand{\segment}{segment}

\newcommand{\model}[1]{\mathcal{M}_\texttt{#1}}
\newcommand{\segModel}{\model{SEG}}
\newcommand{\repairModel}{\model{NOR}}
\newcommand{\interpretModel}[1]{\model{INT(#1)}}
\newcommand{\stateModel}{\interpretModel{state}}
\newcommand{\programModel}{\interpretModel{program}}

\newfloatcommand{capbtabbox}{table}[][\FBwidth]

\newcommand{\makename}[3][s]{%
  \expandafter\newcommand\csname #2\endcsname{#3\xspace}%
  \expandafter\newcommand\csname #2s\endcsname{#3#1\xspace}%
}
\makename{event}{event}
\makename{utt}{segment}
\makename{asr}{ASR result}
\makename{prog}{program}
\makename{statechange}{state update}
\makename{repair}{repair}
\makename{Event}{Event}
\makename{Asr}{ASR result}
\makename{Utt}{Segment}
\makename{Op}{Segment}
\makename{op}{segment}
\makename{Prog}{Program}
\makename{Statechange}{State update}
\makename{Repair}{Repair}
\makename{audio}{speech-audio stream}
\makename{transcript}{speech-audio stream}
\makename{trajectory}{interactive dictation trajectory}

%% file: latex/comments.tex
\newcommand{\Jason}[1]{}

%% file: sections/0-abstract.tex
Voice dictation is an increasingly important text input modality. Existing systems that allow both dictation and editing-by-voice restrict their command language to flat templates invoked by trigger words. In this work, we study the feasibility of allowing users to interrupt their dictation with spoken editing commands in \emph{open-ended} natural language. We introduce a new task and dataset,~\ourdataset, to experiment with such systems. To support this flexibility in real-time, a system must incrementally segment and classify spans of speech as either dictation or command, and interpret the spans that are commands. We experiment with using large pre-trained language models to predict the edited text, or alternatively, to predict a small text-editing program. Experiments show a natural trade-off between model accuracy and latency: a smaller model achieves 28\% single-command interpretation accuracy with $1.3$ seconds of latency, while a larger model achieves 55\% with $7$ seconds of latency.

%% file: sections/1-intro.tex
\section{Introduction}
\label{sec:intro}
\begin{figure}[ht]
    \centering
    \includegraphics[width=\columnwidth,trim={0 9cm 36cm 0},clip]{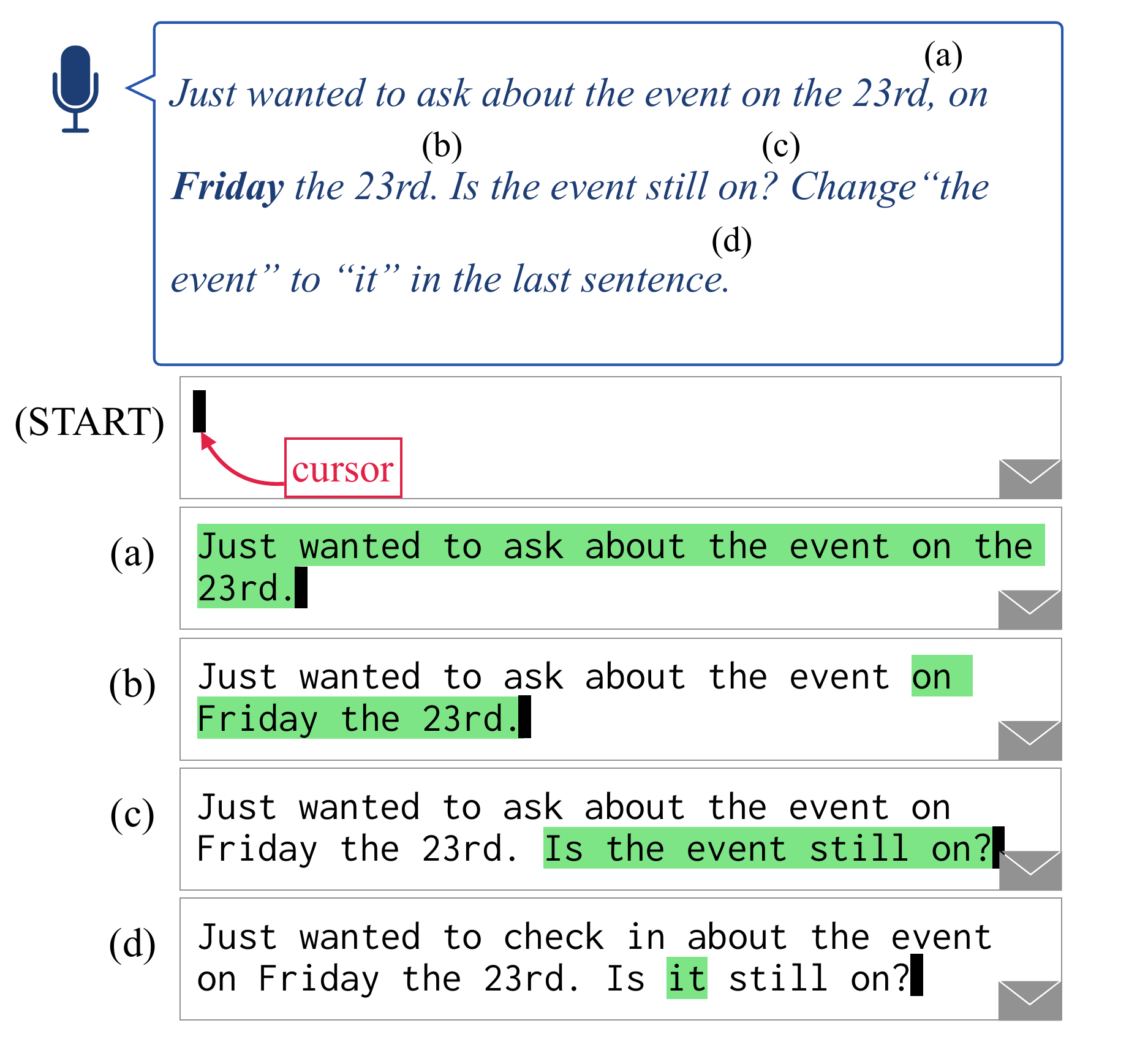}
    \caption{
    A user writes an email using speech input, interleaving dictation (a,c) and commanding (b,d). 
    Top shows the continuous user utterance, while bottom shows the document state at each point of the utterance.
    Dictations are transcribed verbatim, while commands are interpreted and executed.
    Our system supports \textit{open-ended commanding} 
    (\ie b,d both invoke a \texttt{replace} operation but use vastly different phrasing).
    }
    \vspace{-1em}
    \label{fig:teaser}
\end{figure}
Speech can be preferable for text entry, especially on mobile devices or while the user's hands are occupied,
and for some users for whom typing is always slow or impossible.
While fast and accurate automatic speech recognition (ASR) is now ubiquitous~\cite{kumar2012voice,xiong16parity,chiu2017google,whisper},
ASR itself only \emph{transcribes} speech.
In practice, users may also wish to \emph{edit} transcribed text.
The ASR output might be incorrect; 
the user might have misspoken;
or they might change their mind about what to say or how to phrase it, perhaps after seeing or hearing their previous version.
\Citet{Azenkot2013ExploringTU} found that users with visual impairment spent 80\% of time editing text vs.\@ 20\% dictating it.

In this work, we study the task of
\textbf{interactive dictation}, in which users can both perform verbatim dictation and utter open-ended commands in order to edit the existing text, in a single uninterrupted speech stream.
See \cref{fig:teaser} for an example.
Unlike commercial systems like
Dragon~\citep[DNS;][]{DSR,Dragon} and dictation for Word~\citep{DfW} that require reserved trigger words for commanding,
the commands in our data are invoked using unrestricted natural language (NL).
    For example, in~\cref{fig:teaser}, both (b) and (d) invoke \texttt{replace} commands, but (d) uses nested syntax to specify both an edit action and location, while
    (b) is implicit (as natural speech repairs often are).

In interactive dictation, users do not need to memorize a list of specific trigger words or templates in order to invoke their desired functionality.
A dictation system should be as intuitive as dictating to a \emph{human} assistant---a situation in which people quite naturally and successfully intersperse speech repairs and commands with their dictation.  Beyond eliminating the learning curve, letting users speak naturally should also allow them to focus on what they want to say, without being repeatedly distracted by the frustrating separate task of getting those words into the computer. 

Because we accept unrestricted NL for commands, both \emph{segmentation} and \emph{interpretation} become nontrivial for a system to perform.\footnote{%
In template-based systems, by contrast, commands can be detected and parsed using regular expressions.  An utterance is considered a command if and only if it matches one of these regular expressions.
}
Segmentation requires capturing (sometimes subtle) changes in intent, and is especially difficult in cases where command boundaries do not align with ASR boundaries.\footnote{%
In Figure~\ref{fig:teaser}, for example, we must segment the first sentence into two parts, a dictation (\textit{``Just wanted to ask about the event on the 23rd''}) and a command (\textit{``on Friday the 23rd''}).
ASR can also \emph{overpredict} boundaries
when speakers pause in the middle of a sentence.
For example, in our data \textit{``Change elude mansion to elude mentioned.''} was misrecognized by MSS as \textit{``Change. Elude mansion to elude mentioned.''}
}
We collect a 
dataset of 1320 documents dictated in an interactive environment with live, incremental ASR transcription and Wizard-of-Oz--style
interpretation of user commands.
Annotators were not told a set of editing features they were allowed to use, but simply instructed to make their commands understandable and executable by a hypothetical human helper.
Collection required designing a novel data collection interface. Both the interface and
dataset will be publicly released
to help unlock further work in this area.\footnote{\url{https://aka.ms/tertius}}

Finally, we experiment with two strategies for implementing the proposed system: one that uses a pre-trained language model to directly predict the edited text given unedited text and a command, and another that interprets the command as a program specifying how to edit.  Predicting intermediate programs reduces latency because the programs are short, at the expense of accuracy. This strategy also requires additional work
to design and implement a set of editing functions and annotate commands with programs that use these functions.

For each strategy, we also experimented with two choices of pre-trained language model: a small fine-tuned T5 model and a large prompted GPT3 model.
Using the smaller model significantly improves latency, though again at the cost of accuracy.

In summary, our contributions are:
(1) a novel \textit{task} (interactive dictation),
(2) a novel \textit{data collection interface} for the task, with which we collect a new \textit{dataset}, and
(3) a \textit{system} that implements said task, with experiments and analysis.

%% file: sections/2-background.tex
\section{Background \& Related Work}
\label{sec:background}
Many modern speech input tools
only support direct speech-to-text~\cite[\eg][]{whisper}.
Occasionally, these models also perform
disfluency correction, which includes removing filler words (e.g.,
\textit{um}), repeated words, false starts, etc.\@ \citep[\eg][]{MSS}.
One form of disfluency that has received particular attention is speech repair, where the speaker corrects themself
mid-utterance. For example, \textit{let's chat tomorrow uh I mean Friday} contains a speech repair, where the user corrects ``tomorrow'' with ``Friday.''
The repaired
version of this should be \textit{let's chat Friday}.
Prior work has collected datasets and built systems specifically for speech repair~\cite{HeemanAllen94,HeemanAllen99,johnson-charniak-2004-tag}.
Additionally, ASR systems themselves make errors that humans may like to correct post-hoc;
there has been work on correcting ASR errors through \textit{respeaking} misdetected
transcriptions~\cite{mcnair94_icslp,ghosh2020redictation,vertanen_asru2009, sperber2013respeaking}.

Beyond disfluencies that were not automatically repaired but were transcribed literally, humans must fix many other mistakes while dictating.
They often change their mind about what to say---the human writing process is rarely linear---%
and ASR
itself commonly introduces transcription errors.
Most systems require the user to manually fix these errors through keyboard-and-mouse or touchscreen editing~\citep[\eg][]{kumar2012voice}, which can be inconvenient for someone who already relies on voice for dictation.
Furthermore, most commercial systems that support
editing through speech (DNS, Word) require templated commands.
Thus, while speech input is often used to write short-form,
imprecise text (\eg search queries or text messages),
it is not as popular as it might be,
and it is used less when writing longer and more precise documents.

In our work, we study making edits through spoken natural language commands.
Interpreting flexible natural language commands is a well-studied problem within NLP, with work in semantic parsing~\cite{mooney1993,zettlemoyer2009learning,artzi-zettlemoyer-2013-weakly}, instruction-following~\cite{chenmooney,branavan-etal-2009-reinforcement,tellex2011,anderson2018vision,misra-etal-2017-mapping}, and task-oriented dialogue~\cite{budzianowski-etal-2018-multiwoz}.
Virtual assistants like Siri~\cite{siri}, Alexa~\cite{alexa}, and Google Assistant~\cite{googassist} have been built to support a wide range of functionalities, including interacting with smart devices, querying search engines, scheduling events, etc.
Due to advances in language technologies, modern-day assistants can
support flexible linguistic expressions for invoking
commands,
accept feedback and perform reinterpretation~\cite{Andreas:2020:dataflow},
and work in an online and incremental manner~\cite{zhou-etal-2022-online}.
Our work falls in this realm but:
(1) in a novel interactive dictation setting,
(2) with unrestricted commanding, and
(3) where predicting boundaries between dictations and commands is part of the task.

Recently, a line of work has emerged examining how large language models (LLMs) can serve as collaborative writing/coding assistants.
Because of their remarkable ability to generate coherent texts over a wide range of domains and topics, LLMs have proven surprisingly effective for editing, elaboration, infilling, etc.,
across a wide range of domains~\cite{malmi-etal-2022-text,gpt3edit,donahue-etal-2020-enabling}.
Though our system also makes use of LLMs, it supports a different
mode of editing than these prior works.
Some works use edit models for other types of sequence-to-sequence tasks (e.g. summarization, text simplification, style transfer)
~\cite{malmi-etal-2019-encode,dong-etal-2019-editnts,reid-zhong-2021-lewis},
while others use much coarser-grained editing commands than we do, expecting the LLM to (sometimes) generate
new text~\cite{gpt3edit,coditt5}.
In addition to these differences, 
our editing commands may be misrecognized because they are spoken, and may be misdetected/missegmented because they are provided through the same channel as text entry.

%% file: sections/3-task.tex
\begin{figure*}[ht]
    \centering
    \includegraphics[width=\textwidth,trim={0 9.5cm 0 0},clip]{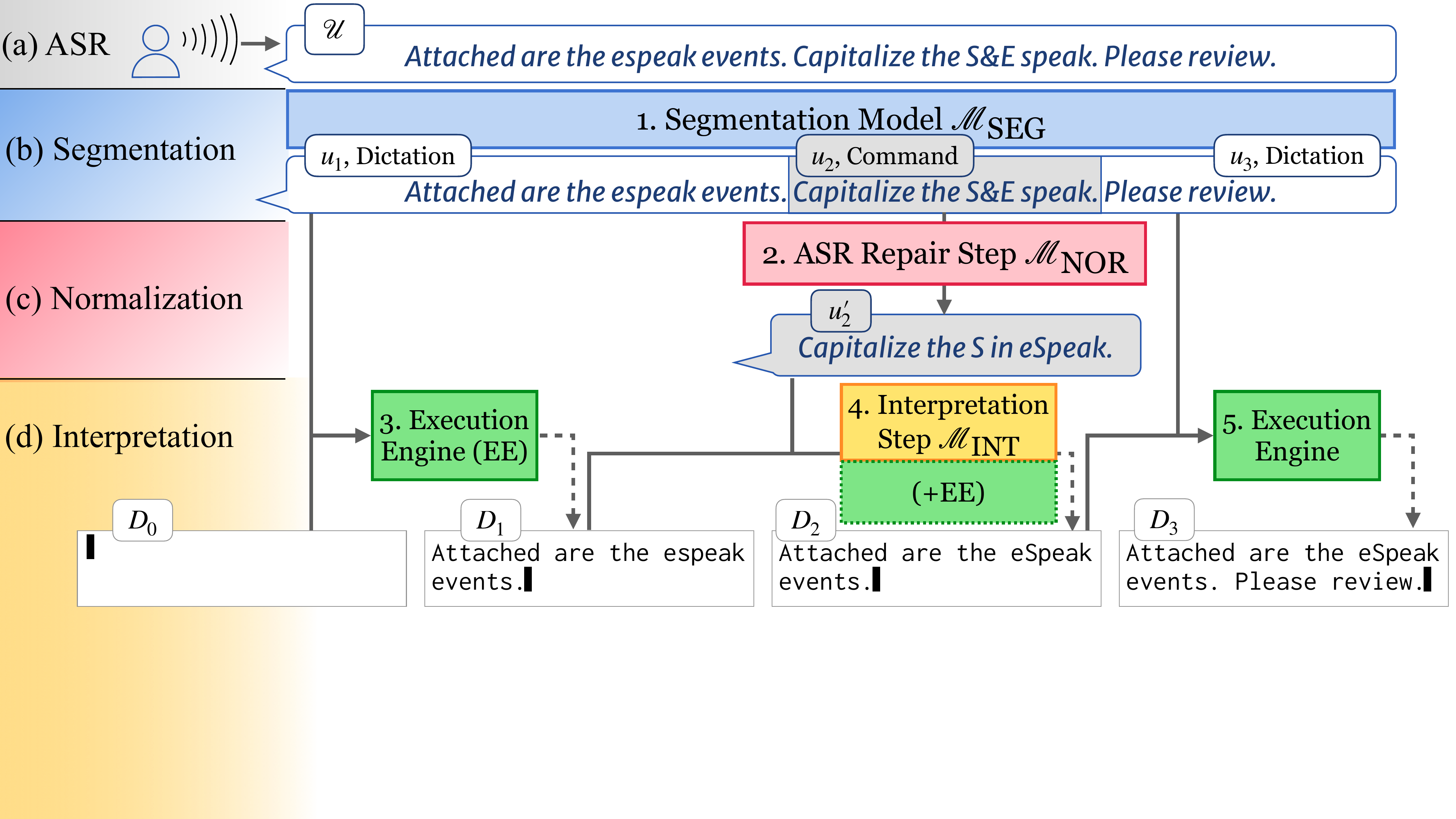}
    \caption{%
    Diagram of an interactive dictation system.
    First, the ASR system (a) transcribes speech, which the segmentation system (b) parses into separate dictation and command segments. 
    Next, an optional normalization module (c) fixes the any ASR or speech errors in the segment.
    Finally, the interpretation system (d) returns the result of each operation.
    On the right is the concrete instantiation of \textit{our} system.
    }
    \label{fig:model}
\end{figure*}

\section{Task Framework}
\label{sec:preliminaries}

We now formalize our \taskname{} setting.
A user who is editing a document speaks to a system that both transcribes user dictation and responds to user commands.  This process results in a \defn{\trajectory}%
---a sequence of timestamped events: the user keeps speaking, several trained modules keep making predictions, and the document keeps being updated.  

Supervision could be provided to the predictive modules in various ways, ranging from direct supervision to delayed indirect reward signals.  In this paper, we collect supervision that can be used to bootstrap an initial system.
We collect
\defn{gold trajectories} in which every prediction is correct---except for ASR predictions, where we preserve the errors since part of our motivation is to allow the user to fix dictation errors.\footnote{In module (c) below, we predicted repairs for command segments, so the gold trajectory interprets accurate clean text for commands.  But we did not predict repairs for dictation segments, so their errors persist even in the gold trajectories.}  
All predictions along the trajectory are provided in the dataset.

Our dataset is not completely generic, since it assumes that certain predictive modules will exist and interact in particular ways, although it is agnostic to how they make their predictions.  It is specifically intended to train a system that is a pipeline of the following modules (\cref{fig:model}): 

\noindent\textbf{(a) ASR}\quad As the user speaks, the ASR module proposes transcripts for spans of the audio stream. 
Due to ASR system latency, each \asr normally arrives some time \emph{after} the end of the span it describes.
The \asrs are transcripts of successive disjoint spans of the audio, and we refer to their concatenation as the \defn{current transcript} ($\allUttsConcated$ in~\cref{fig:model}(a)).  

\noindent\textbf{(b) Segmentation}\quad
When the current transcript changes, the system can update its segmentation.  It does so by partitioning the current transcript $\allUttsConcated$ into a sequence of segments $u_i$, labeling each as being either a \defn{dictation} or a \defn{command}.  

\noindent\textbf{(c) Normalization \emph{(optional)}}\quad Each segment $u_i$ can be passed through a normalization module, which transforms it from a literal transcript into clean text that should be inserted or interpreted.
This involves speech repair as well as text normalization to handle orthographic conventions such as acronyms, punctuation, and numerals.  

While the module (a) may already attempt some version of these transformations, an off-the-shelf ASR module
does not have access to the document state or history.
It may do an incomplete job and there may be no way to tune it on gold normalized results.
This normalization module can be trained to finish the job.
Including it also ensures that our gold trajectories include the intended normalized text of the commands.

\noindent\textbf{(d) Interpretation}\quad Given a document state $d_{i-1}$ and a segment $u_i$, the interpretation module predicts the new document state $d_i$ that $u_i$ is meant to achieve.\footnote{This prediction can also condition on earlier segments, which provide some context for interpreting $u_i$.
It might also depend on document states other than $d_{i-1}$---such as the state or states that were visible to the user while the user was actually uttering $u_i$,
for example.  
}
The document is then immediately updated to state $d_i$; the change could be temporarily highlighted for the user to inspect.  Here $d_{i-1}$ is the result of having already applied the updates predicted for segments $u_1, \ldots, u_{i-1}$, where $d_0$ is the initial document state.  
Concretely, we take a \defn{document state} to consist of the document content together with the current cursor position.\footnote{The cursor may have different start and end positions if a span of text is selected, but otherwise has width 0.  For example, the document state $d_1$ in~\cref{fig:model} is $(\textit{"Attached are the espeak events."},$ $(31, 31))$.}

When $u_i$ is a dictation segment, no prediction is needed: the \statechange simply inserts the current segment at the cursor.  However, when $u_i$ is a command segment, predicting the \statechange that the user wanted requires a text understanding model.  Note that commands can come in many forms.  Commonly they are imperative commands, as in \cref{fig:teaser}d.  But one can even treat speech repairs such as \cref{fig:teaser}b as commands, in a system that does not handle repairs at stage (a) or (c).

Rather than predict $d_i$ directly, an alternative design is to predict a program $p_i$ and apply it to $d_{i-1}$ to obtain $d_i$.  
In this case, the gold trajectory in our dataset includes a correct program $p_i$, which represents the intensional semantics of the command $u_i$ (and could be applied to different document states).

\paragraph{Change Propagation}

The ASR engine we use for module (a) sometimes revises its results.  It may replace the most recent of the \asrs, adding new words that the user has spoken and/or improving the transcription of earlier words.  The engine marks an \asr as \defn{partial} or \defn{final} according to whether it will be replaced.\footnote{Full details and examples can be found in~\cref{sec:app_ASR}.}

To make use of streaming partial and final ASR results, our pipeline supports change propagation.  This requires the predictive modules to compute additional predictions.  If a module is notified that its input has changed, it recomputes its output accordingly.  For example, if module (a) changes the current transcript, then module (b) may change the segmentation.  Then module (c) may recompute normalized versions of segments that have changed.  Finally, module (d) may recompute the document state $d_i$ for all $i$ such that $d_{i-1}$ or $u_i$ has changed.  

The visible document is always synced
with the last document state. This sync can revert and replace the effects on the document of previous incorrectly handled dictations and commands, potentially even from much earlier segments.  To avoid confusing the user with such changes, and to reduce computation, a module can freeze its older or more confident inputs so that they reject change notifications (\cref{sec:commitpoint}).
Modules (b)--(d) could also adopt the strategy of module (a)---quickly return provisional results from a ``first-pass'' system with the freedom to revise them later.
This could further improve the responsiveness of the
experience.

%% file: sections/4-dataset.tex
\section{Dataset Creation}
\label{sec:dataset}
\begin{figure*}[ht]
    \centering
    \includegraphics[width=\textwidth,trim={5cm 5cm 0 0},clip]{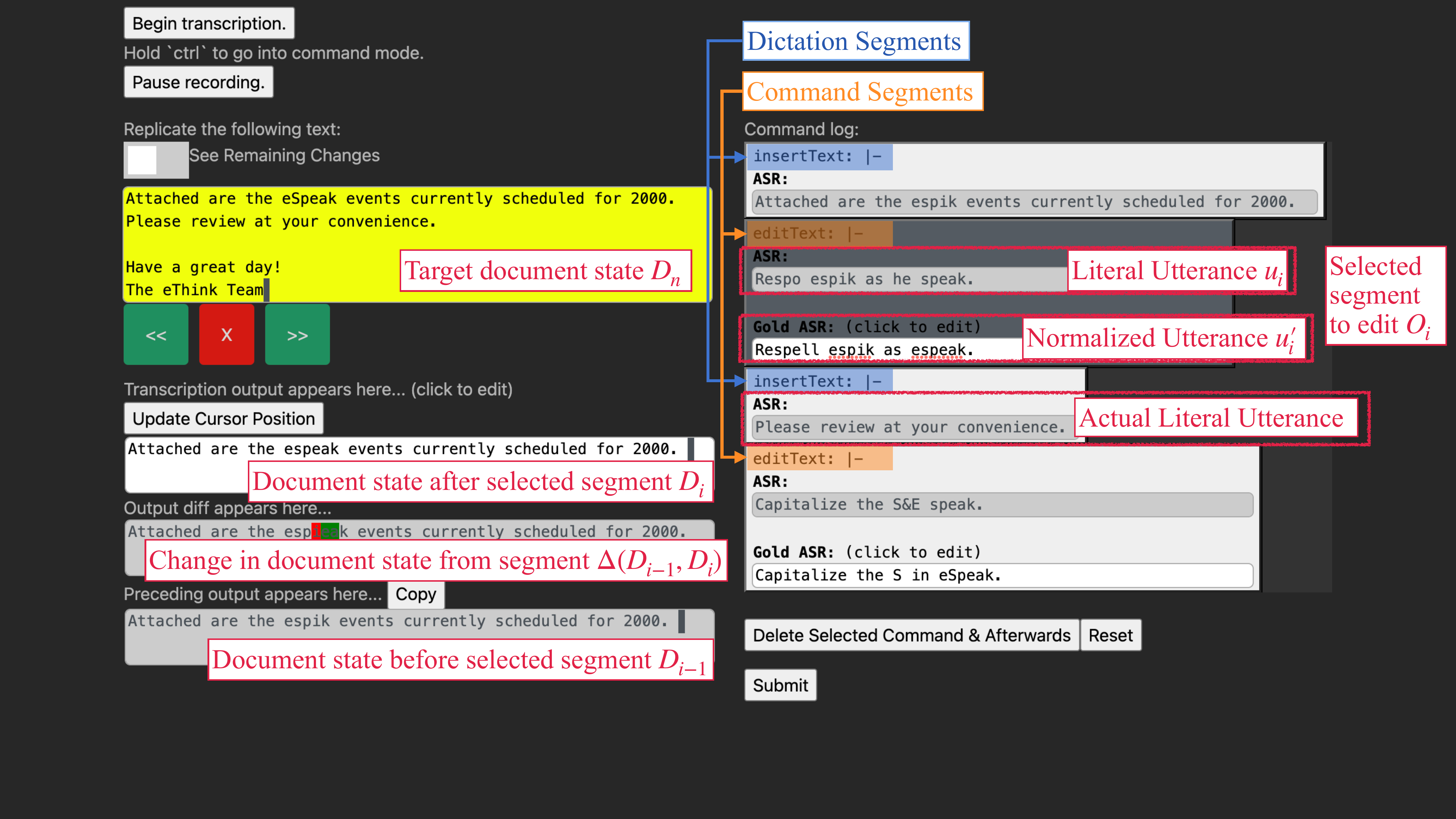}
    \caption{Data collection UI.
    Demonstrator speech is transcribed by a built-in ASR system.
    Demonstrators specify gold segmentations by pressing a key to initiate a command \op (\texttt{editText}) and releasing the key to initiate a dictation \op (\texttt{insertText}).
    The resulting transcribed segments appear in the \texttt{ASR} fields of the boxes in the right column.
    For a command \op, the demonstrator specifies the normalized version in the 
    \texttt{Gold ASR} field,
    and demonstrates the command interpretation by editing the document post-state.
    Document states are shown in the left column: selecting a \op makes its post-state (and pre-state) appear there.
    }
    \label{fig:UI_screenshot}
\end{figure*}
To our knowledge, no public dataset exists for the task of interactive dictation.
As our task is distinct from prior work in a number of fundamental ways~(\cref{sec:background}),
we create a new dataset,~\ourdataset.\footnote{%
\textbf{T}ranscribing and \textbf{E}diting in \textbf{R}eal-\textbf{Ti}me with \textbf{U}nrestricted \textbf{S}peech.
Named for the human amanuensis \href{https://en.wikipedia.org/wiki/Tertius_of_Iconium}{Tertius of Iconium}.}

Our data collection involves two stages.
First, a human \defn{demonstrator}
speaks to the system and provides the gold segmentations, as well as demonstrating the normalizations and document state updates for the command \ops.
Later, for each command \op, an \defn{annotator}
fills in a gold \prog that would yield its gold \statechange.

For a command \ops, we update the document during demonstration using the demonstrated \statechanges---%
that is, they do double duty as \emph{gold} and \emph{actual} \statechanges.  Thus, we follow a gold trajectory, as if the demonstrator is using an oracle system that perfectly segments their speech into dictations (though these may have ASR errors) versus commands, and then perfectly interprets the commands.
A future data collection effort could instead update the document using the imperfect system that we later built (\cref{sec:model}), in which case the demonstrator would have to react to cascading errors.\looseness=-1

\subsection{Collecting Interactive Dictation}
\label{sec:dialogue_UI}
We build a novel data collection framework that allows us to
collect speech streams and record gold and actual \events.

We used an existing ASR system, Microsoft Speech Services \citep[\defn{MSS};][]{MSS}.
We asked the demonstrator to play both the role of the \textit{user} (issuing the speech stream), and also the roles of the \textit{segmentation}, \textit{normalization}, and \textit{interpretation} parts of the system (Figures \labelcref{fig:model}b--d).
Thus, we collect actual \asrs, while asking the demonstrator to demonstrate gold predictions for segmentation, normalization, and interpretation.

The demonstration interface is shown in ~\cref{fig:UI_screenshot}.
demonstrators were trained to use the interface, and told during training how their data would be used.\footnote{%
We met with demonstrators ahead of time and provided them with written instructions, which 
are in~\cref{sec:app_annot_instructions}.}
A demonstrator is given a task of dictating an email
into our envisioned system (shown in the yellow textbox).
We collected data in three scenarios:
\begin{enumerate}
\item    
\textbf{\enronTask:}
Exactly recreate an email from the Enron Email Dataset~\cite{enron}%
.\footnote{Preprocessing details can be found in~\cref{sec:app_targets}.}
\item
\textbf{\emailExpandTask:} Expand a terse description of an email into an full email.
The exact wording of the full email is up to the demonstrator. 
\item
\textbf{\replicateOpTask:} Exactly recreate the post-state $d_i$ of a single command \op $u_i$ (randomly sampled from the already-collected \enronTask\ and \emailExpandTask\ data), starting from its pre-state $d_{i-1}$.  This does not have to be done with a single command.
\end{enumerate}
A demonstrator must then reach the target state (either exactly for \enronTask\ or \replicateOpTask, or to their satisfaction for \emailExpandTask), following these three steps:

\noindent\textbf{Step 1 (ASR, segmentation)}\quad The demonstrator starts speaking, which gets transcribed in real time by the built-in ASR system into \asrs.
As they speak, they demonstrate what the segmentation system should do by holding down a key whenever they are speaking a command (as opposed to dictating).
They can specify consecutive commands by quickly releasing and re-pressing the key.\footnote{We do not allow two dictation segments to be adjacent---that would be equivalent to one concatenated segment.}
This gives us a list 
of time intervals when the key was held down.
By matching
these to the ASR timestamps, we identify the gold command segments in the ASR transcript.
The remaining \ops of the transcript are labeled as dictation.%
\footnote{More details on how the \asrs are combined/segmented can be found in~\cref{sec:app_ASR}.}

\noindent\textbf{Step 2 (normalization)}\quad All labeled \ops are displayed in the right column of the UI. 
After the demonstrator has finished speaking, they fill in the normalized text for each command \op.
(The segment shows original and normalized text in the \texttt{ASR} and \texttt{Gold ASR} fields.)

\noindent\textbf{Step 3 (interpretation)}\quad Finally, for each command \op, the demonstrator manually carries out the gold \statechange.\footnote{For dictation \ops, the system automatically computes the gold \statechange by inserting the \utt at the selection. This \utt is an actual \asr and may contain errors.}
They do this by clicking on a command \op $u_i$ in the right column, which pulls up the associated document state $d_i$ in the left column.  Initially $d_i$ is set to equal the pre-state $d_{i-1}$, and the demonstrator edits it with their mouse and keyboard until it reflects the desired post-state after applying command $u_i$.  For reference, the UI also displays the pre-state $d_{i-1}$ and a continuously updated visual diff $\Delta(d_{i-1}, d_i)$).

Demonstrators can move freely among these steps, editing normalizations or \statechanges at any time, or appending new segments by speaking.\footnote{They are also allowed to back up and remove the final segments, typically in order to redo them.} 

We believe our
framework is well-equipped to collect natural, flexible, and intuitive
dictation and commanding data, for several reasons:
(1) We do not restrict the capabilities of commands or the forms of their utterances, but instead
ask demonstrators to command in ways they find most natural.
(2) We simulate natural, uninterrupted switching between \ops by making it easy for demonstrators to specify \op boundaries in real-time.
(3) We collect a realistic distribution over speech errors and corrections by using an existing ASR system and asking demonstrators to replicate real emails.  In the future, the distribution could be made more realistic if we sometimes updated the document by using predicted normalizations and state updates rather than gold ones, as in the DAgger imitation learning method \cite{ross11dagger}.

\subsection{Annotating Programs for Commands}
\label{sec:program}
After obtaining sequences of demonstrated dialogues
using the above procedure, we extract each \textit{command} \op and manually annotate it with a \textbf{\prog} $p_i$ that represents the intensional semantics of the command. This \prog should in theory output the correct $d_i$ when given $d_{i-1}$ as input.
\Prog annotation is done post-hoc with a different set of annotators from~\cref{sec:dialogue_UI}.

We design a domain-specific Lisp-like language for
text-manipulating programs, and an execution engine for it.
We implement a library consisting of composable \emph{actions},
\emph{constraints},
and \emph{combinators}.
A \prog consists of actions
applied to one or more text targets, which are specified by contraints.
Combinators
allow us to create
complex constraints by composing them.
For example,
in~\cref{fig:model}, \textit{Capitalize the S in eSpeak}, has the program

\begin{small}
\begin{verbatim}
(capitalize
  (theText
    (and
      (like "S")
      (in (theText (like "eSpeak"))))))
\end{verbatim}
\end{small}
where \verb|capitalize| is the action, \verb|(like "S")| and \verb|(like "eSpeak")| are constraints, and \verb|and| is a
combinator.
More examples are in~\cref{sec:app_lispress}.

\subsection{Handling of partial \asrs}

The current transcript sometimes ends in a partial \asr and then is revised to end in another partial \asr or a final \asr.  All versions of this transcript---``partial'' and ``final''---will be passed to the segmenter, thanks to change propagation.  During demonstration, we record the gold labeled segmentations for all versions, based on the timing of the demonstrator's keypresses.  

However, only the segments of the ``final'' version are shown to the demonstrator for further annotation.  A segment of a ``partial'' version can simply copy its gold normalized text from the segment of the ``final'' version that starts at the same time.  These gold data will allow us to train the normalization model to predict a normalized command based on partial \asrs, when the user has not yet finished speaking the command or the ASR engine has not yet finished recognizing it.

In the same way, a command segment $u_i$ of the ``partial'' version could also copy its gold document post-state $d_i$ and its gold program $p_i$ from the corresponding ``final'' segment.  However, that would simply duplicate existing gold data for training the interpretation module, so we do not include gold versions of these predictions in our dataset.\footnote{The gold pre-state $d_{i-1}$ may occasionally be different, owing to differences between the two versions in earlier dictation segments.  In this case, the interpretation example would no longer be duplicative (because it has a different input).  Unfortunately, in this case it is no longer necessarily correct to copy the post-state $d_i$, since some differences between the two versions in the pre-state might need to be preserved in the post-state.}

\subsection{Dataset details \& statistics}
\label{sec:dataset_details}

\begin{table}
    \centering
    \footnotesize
    \begin{tabular}{@{}l r c rrr@{}}
    \toprule
        & \textbf{Trajectories} & & \multicolumn{3}{c}{\textbf{\Ops}} \\
        \textbf{Task} & & & \small{Dict.} & \small{Cmd.} & \small{Total} \\
    \midrule
        \enronTask & $372$ && $473$ & $1453$ & $1926$ \\
        \emailExpandTask & $343$ && $347$ & $473$ & $820$ \\
        \replicateOpTask & $605$ && $139$ & $1299$ & $1438$ \\
\midrule
        \textbf{Total} & $1320$ && $959$ & $3225$ & $4184$ \\
    \bottomrule
    \end{tabular}
    \caption{Dataset size statistics.}
    \label{tab:data}
\end{table}

In the first stage
(\cref{sec:dialogue_UI}),
eleven human demonstrators demonstrated 1372 interactive dictation trajectories (see~\cref{tab:data} for details).
In the second stage
(\cref{sec:program}), two human annotators annotated programs for 868 commands.\footnote{The rest of the programs were auto-generated by GPT3. See details in~\cref{sec:app_interpret_training}.}
The dataset was then split into training, validation, and test sets with 991 training trajectories (consisting of 3199 demonstrated \ops), 173 validation trajectories (562 \ops), and 156 test trajectories (423 \ops).

All demonstrators and annotators were native English speakers. 
The dataset is currently only English, and
the editor supports unformatted plain text.
However, the annotation
framework could handle other languages that have spoken and written forms, and could be extended to allow formatted text.

A key goal of our system is flexibility.
We quantify how well~\ourdataset captures
flexibility by measuring 
the \textit{diversity} of natural language used to invoke each state change.\footnote{The system we build can theoretically support more flexibility than what is captured in~\ourdataset. However, for~\ourdataset to be a
useful testbed (and training set) for flexibility, we would like it to be itself diverse.}
We count the number of distinct first tokens (mainly verbs) used to invoke each action. These results are reported in~\cref{tab:data_invoke_diversity} in the Appendix, alongside a comparison with DNS.\footnote{We also measure the diversity of state changes captured by~\ourdataset in~\cref{sec:app_data_breakdown}.}
We see that~\ourdataset contains at least 22 ways to invoke a \verb|correction|, while DNS supports only 1. 
In short, these results show that doing well on~\ourdataset requires a much more flexible system that supports a wider array of functions and ways of invoking those functions than what existing systems provide.

%% file: sections/5-models.tex
\section{Modeling \& Training}
\label{sec:model}
The overall system we build for \taskname\ 
follows our pipeline from \cref{fig:model,sec:preliminaries}:
\begin{enumerate}
    \item A \textbf{segmentation model} $\segModel$ takes the current transcript 
    $\allUttsConcated$,
    and predicts a segmentation $u_1,\ldots,u_n$,
    simultaneously predicting whether each $u_i$ corresponds to a \textit{dictation} or \textit{command} \op.
    \item Each dictation \op is directly spliced into the document at the current cursor position.
    \item For each command \op: 
    \begin{enumerate}
        \item A \textbf{normalization model} $\repairModel$
        predicts the normalized utterance $u_i'$, repairing any
        ASR misdetections.

        \item An \textbf{interpretation model}, $\stateModel$ or $\programModel$,
        either: 1. directly predicts the end state of the command $d_i$, or 2. predicts the command \textit{program} $p_i$, which is then executed to $d_i$ by the \textbf{execution engine}.
        We experiment with both types of interpretation model.
    \end{enumerate}
\end{enumerate}
Below we describe the specific models we use.

\subsection{Segmentation}
\label{sec:segmentation_model}
The segmentation model partitions $\allUttsConcated$ into segments $u_i$, each of which is labeled by $m_i$ as being either dictation or command:
\begin{align}
\begin{split}
    \mathcal{M}_\texttt{SEG}(\allUttsConcated) &= [(u_0, m_0),\cdots, (u_n, m_n)], \\
    \text{ s.t. } \allUttsConcated &= u_0 + u_1 + \cdots + u_n \\ 
    m_i &\in \{\text{command}, \text{dictation}\}
\end{split}
\end{align}
Concretely, the segmentation model does this using
BIOES tagging~\cite[Chapter~5]{jurafsky2009speech}.
Here each command is tagged
with a sequence of the form \texttt{BI$^*$E} (``beginning, inside, \ldots, inside, end'') or with the length-1 sequence \texttt{S} (``singleton'').  Maximal sequences of tokens tagged with \texttt{O} (``outside'') then correspond to the dictation segments.  Note that two dictation segments cannot be adjacent.
We implement the segmentation model as a T5-base encoder~\cite{t5} followed by a two-layer MLP prediction module. 
More details on why each tag is necessary and how we trained this model can be found in~\cref{sec:app_seg_training}. 

\subsection{Normalization and Interpretation}

For each $u_i$ that is predicted as
a command \op, we first predict the
normalized utterance $u_i'$,\footnote{Note that the normalization step additionally conditions on the state $d_{i-1}$, allowing it to consider what command would have been sensible in this context. Concrete examples are given in~\cref{sec:app_eval_asr}.}
\begin{align}
\repairModel (d_{i-1}, u_i) = u_i'.
\end{align}
We then interpret $u_i'$ in context to 
predict either the document state 
$d_i$ or an update program $p_i$.
\begin{align}
\begin{split}
\stateModel (d_{i-1},u_i') &= d_i, \\
\programModel (d_{i-1},u_i') &= p_i.
\end{split}
\end{align}
We then update the document state accordingly.

We experiment with two ways of implementing the two steps:
we either fine-tune two separate T5-base models~\cite{t5} that run in a pipeline for each command, or we prompt GPT3~\cite{GPT3}\footnote{Specifically, the \texttt{text-davinci-003} model.} to generate both the normalized utterance\footnote{Although the normalized utterance is not used for the final state prediction, early experiments indicated that this auxiliary task helped the model with state prediction, possibly due to 
 a chain-of-thought effect~\cite{wei2022chainofthought}.} and the interpretation output in a single inference step.
Training and prompting details
can be found in~\cref{sec:app_interpret_training}.

%% file: sections/6-results.tex
\begin{figure*}[t]
\begin{floatrow}
\capbtabbox{%
    \small
    \begin{tabular}{@{}llcccc@{}}
    \toprule
         & \textbf{Metric} & \multicolumn{2}{c}{\textbf{T5}} & \multicolumn{2}{c}{\textbf{GPT3}} \\
        \midrule
        \multirow{3}{*}{Segmentation} & F1 & \multicolumn{2}{c}{$90.9\%$} & \multicolumn{2}{c}{-} \\ 
        & Segmentation EM & \multicolumn{2}{c}{$85.3\%$} & \multicolumn{2}{c}{-} \\ 
        & Runtime (s/it) & \multicolumn{2}{c}{$0.097$} & \multicolumn{2}{c}{-} \\
        \midrule
        & & \textbf{prog} & \textbf{state} & \textbf{prog} & \textbf{state} \\
        \cmidrule(lr){3-3}\cmidrule(lr){4-4}\cmidrule(lr){5-5}\cmidrule(lr){6-6}
        ASR Repair $+$ & State EM & $28.3\%$ & $29.5\%$ & $38.6\%$ & $55.1\%$ \\ 
        Interpretation & Program EM & $28.3\%$ & - & $41.9\%$ & - \\ 
        & Runtime (s/it) & $1.28$ & $3.46$ & $5.32$ & $6.92$ \\
        \bottomrule
    \end{tabular}
}{%
  \caption{We evaluate
    segmentation (top) and
    the ASR repair and interpretation components jointly (bottom). 
    We report accuracy metrics (F1, EM) as well as runtime (in seconds per example).
    Segmentation
    is relatively fast and performs decently.
    For
    ASR repair and interpretation, 
    we experiment with a fine-tuned T5 vs. a prompted GPT3 model, each outputting either the end state (state) or a program to carry out the command (prog). 
    }%
    \label{tab:results}
}
\ffigbox[0.35\textwidth]{%
  \includegraphics[scale=0.3]{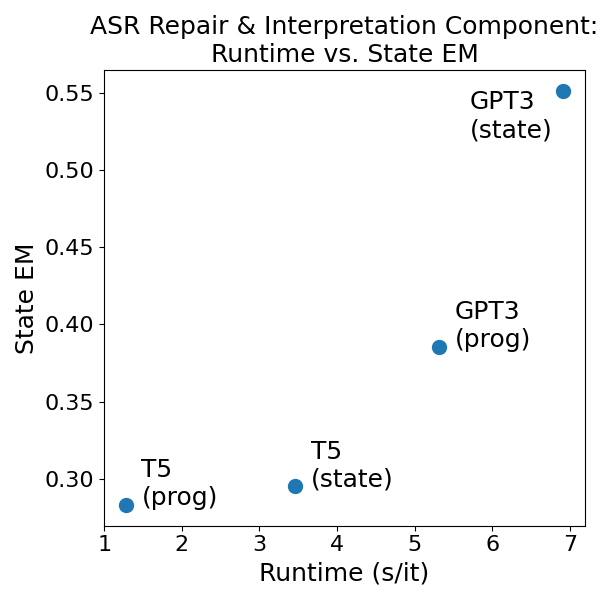}
}{%
    \caption{
    Runtime vs. State EM.
    GPT3 models produce more accurate state updates than
    T5,
    but use an unreasonable amount of time. 
    Directly predicting the updated documents is more often correct than predicting update programs, again at the cost of time.
    }
    \label{fig:tradeoff}
}
\end{floatrow}
\end{figure*}

\section{Results}
\label{sec:results}
We evaluate the segmentation model in isolation, and the normalization and interpretation steps together. (\Cref{sec:app_eval_asr,sec:app_eval_interpret} evaluate the normalization and interpretation steps in isolation.)

For simplicity, we evaluate the models only on current transcripts $\allUttsConcated$ that end in \textbf{final} \asrs (though at training time and in actual usage, they also process transcripts that end in \textbf{partial} ones).\footnote{See~\cref{sec:app_results} for details.}

\subsection{Segmentation}
\paragraph{Metrics}
\textbf{Exact match (EM)} returns 0 or 1 according to whether the entire labeled segmentation of the final transcript $\allUttsConcated$ is correct.
We also evaluate macro-averaged \textbf{labeled F1},
which considers how many of the gold labeled segments appear in the model's output segmentation and vice versa.  Two labeled segments are considered to be the same if they have the same start and end points in $\allUttsConcated$ and the same label (dictation or command).  

\paragraph{Results}
Segmentation results on an evaluation dataset of transcripts $\allUttsConcated$ (see \cref{sec:app_results_segmentation}) are shown in the top section of~\cref{tab:results}.
All results are from single runs of the model.
The model performs decently on~\ourdataset, and in some cases is even able to fix erroneous sentence boundaries detected by the base ASR system (\cref{sec:app_seg_successes}).
However, these cases are also difficult for the model: 
a qualitative analysis of errors find that, generally, errors arise either when the model is misled by erroneous over- and under-segmentation by the base ASR system, or when commands are phrased in ways similar to dictation.
Examples are in
in \cref{sec:app_seg_errors}. 

\subsection{Normalization \& Interpretation}
\label{sec:eval_repair_interpret}
\paragraph{Metrics}
We evaluate normalization and interpretation in conjunction. 
Given a gold normalized command utterance $u_i$
and the document's gold pre-state $d_{i-1}$, we measure how well we can reconstruct its post-state $d_i$.
We measure \textbf{state exact match (EM)}\footnote{We disregard the cursor position in this evaluation.}
between the predicted and gold post-states. 
If the interpretation model predicts intermediate programs, then we also measure \textbf{program exact match (EM)} between the predicted program and the gold program.

\paragraph{Results}
The bottom
of \cref{tab:results} shows these results.
All results are from single runs of the model.
GPT3 generally outperforms T5, likely due to its larger-scale pretraining.
When we evaluated ASR repair and interpretation separately in~\cref{sec:app_eval_asr,sec:app_eval_interpret}, we found that GPT3 was better than T5 at both ASR repair and interpretation.

Furthermore, we find that \textit{both GPT3 and T5 are better at directly generating states} ($55.1$ vs. $38.6$ state EM and $29.5$ vs. $28.3$ state EM). However, the gap is larger for GPT3.
We suspect that
GPT3
has a better prior over well-formed English text and can more easily generate edited documents $d$ directly, without needing the  abstraction of an intermediate program.
T5-base, on the other hand,
finds it easier to learn the distinctive 
(and more direct) relationship between
$u$ and the short program $p$. 

Other than downstream data distribution shift, we hypothesize that program accuracy is lower than state accuracy because the interpretation model is trained mostly on \textit{auto-generated} program annotations, and because the execution engine is imperfect. We anticipate that program accuracy would improve with more gold program annotations and a better execution engine.

\subsection{Efficiency}
\cref{tab:results} reports runtimes for each component.
This allows us to identify bottlenecks in the system and consider trade-offs between model performance and efficiency.
We see that segmentation is generally quick and the ASR repair and interpretation steps are the main bottlenecks. 
The T5 model also runs much faster than the GPT3 model,\footnote{Note that GPT3 is called via an external API, while T5 is run on a local GPU. GPT3 runtimes thus include an unknown communication overhead, which will not be present when run on local hardware.} despite performing significantly worse, indicating a trade-off
between speed and accuracy. 

\cref{fig:tradeoff} shows that by generating programs instead of states, we achieve faster runtimes (as the programs are shorter), at the expense of accuracy.

%% file: sections/7-conclusion.tex
\section{Conclusion}
Most current speech input systems do not support voice editing.
Those that do usually only support a narrow set of commands specified through a fixed vocabulary.
We introduce a new task for \textit{flexible invocation} of commands through natural language, which \textit{may be interleaved with dictation}.
Solving this task requires both \textit{segmenting} and \textit{interpreting} commands.
We introduce a novel data collection
framework that allows us to collect a pilot dataset,~\ourdataset, for this task.
We 
explore tradeoffs between model accuracy and efficiency.
Future work can examine techniques to push out the Pareto frontier, such as model distillation to improve speed and training on larger
datasets to improve accuracy.
Future work can also look at domains outside of (work) emails, integrate other types of text transformation commands (\eg formatting), and may allow the system to respond to the user in ways beyond updating the document.

%% file: sections/999-appendix.tex
\section{Dataset}
\label{sec:app_dataset}
\subsection{ASR results}
\label{sec:app_ASR}

\paragraph{Types of \utts}
Below we describe the types of ASR results we collect in~\ourdataset. 
As dialogues are uttered, we obtain a stream of timestamped partial and full ASR results from MSS.
Examples of partial and full ASR results can be found below:

\begingroup
\begin{quote}
    0:00.00: \textit{attached} \\
    0:00.30: \textit{attached is} \\
    0:00.60: \textit{attached is the} \\
    0:01.05: \textit{attached is the draft} \\
    0:02.15: \textit{Attached is the draft.}
\end{quote}
\label{quote:sample_asr_results}
\endgroup
The first four lines are partial ASR results $u^\text{partial}$ that are computed quickly and returned 
by MSS in real time as the user is speaking. The last line is the final ASR result, which takes slightly longer to compute, but represents a more reliable and polished ASR result.
After a final result $u^\text{final}$ has been computed, it obsolesces prior partial ASR results.

While not used in present experiments, collecting partial ASR results enables building an incremental system that can be faster and more responsive in real time;
rather than waiting for ends of sentences to execute commands, a system can rely on partial ASRs to anticipate commands ahead of time (akin to~\citet{zhou-etal-2022-online}).
Collecting timing information is also helpful for evaluating the speed of our system: the system runtime continges on the rate at which it obtains new ASR results and how long it takes to process them.

Furthermore, MSS additionally returns $n$-best lists for each final ASR results. These are a list of candidate final ASRs that may feasibly correspond with user audio, \eg
\begin{quote}
    \textit{Attached is the draft.} \\
    \textit{Attached his draft.} \\
    \textit{Attacked is the draft.} \\
    $\cdots$
\end{quote}

\paragraph{Aggregation~\utts}
For long user audio streams, partial and final results are returned sequentially, each describing roughly a single sentence.
The most recent ASR result is concatenated together with the previous history of final ASR results, to return the full partial or final ASR result for the entire stream.
For example, after the user utters the first sentence in the example above, the user may continue by saying:
\begin{quote}
    \textit{please} \\
    \textit{please re} \\
    \textit{please review} \\
    \textit{please review win} \\
    \textit{please review when pause} \\
    \textit{please review when possible} \\
    \textit{Please review when possible.}
\end{quote}
We concatenate each of these new ASR results with the previous final ASR results to obtain the current transcript $\allUttsConcated$ (see \cref{sec:preliminaries}),%
which evolves over time as follows:
\begin{quote}
    \textit{Attached is the draft. please} \\
    \textit{Attached is the draft. please re} \\
    \textit{Attached is the draft. please review} \\
    \textit{Attached is the draft. please review win} \\
    \textit{Attached is the draft. please review when pause} \\
    \textit{Attached is the draft. please review when possible} \\
    \textit{Attached is the draft. Please review when possible.}
\end{quote}

\paragraph{Segmenting \Asrs into \Ops During Annotation}
During annotation (\cref{sec:dialogue_UI}), all these partial and final \asrs get mapped to \ops, forming $u_{i}^\text{final}$ and $u_{i}^\text{partial}$. This is done by identifying the \textit{timestamp of each token} within each partial and 
final result.
For example, in the example \asrs sequence at the beginning of this section~\ref{quote:sample_asr_results}, suppose the user specifies an \op boundary at time 0:00.45, (separating ``\textit{Attached is}'' from ``\textit{the draft.}''). We get the following \asrs for the first \op:
\begin{quote}
    \textit{attached} \\
    \textit{attached is} \\
    \textit{Attached is}
\end{quote}
(we refer to the first two as partial ASRs for the \op, as they are derived from partial ASR, and the third as the final ASR for the \op), and the following \asrs for the second \op:
\begin{quote}
    \textit{the} \\
    \textit{the draft} \\
    \textit{the draft.}
\end{quote}

\subsection{Annotation Instructions (\cref{sec:dialogue_UI})}
\label{sec:app_annot_instructions}
The full text of written instructions given to annotators during the first round of annotation (\cref{sec:dialogue_UI}) is provided below:
\begin{enumerate}
\item \textbf{Transcribing}

Your goal is to replicate the prompt in the target box verbatim / expand the prompt in the yellow textbox into a coherent email, starting from the given (potentially non-empty) starting document in the `Transcription output` box. You are expected to do so using a series of speech-to-text transcriptions and commands. Try to use the starting document as much as possible (i.e. do not delete the entire document and start over).

You can easily see what changes are to be made by toggling the `See Diff View` button. Once that mode is on, the text you need to add will be highlighted in green, while the text you need to delete will by highlighted in red. Once there is no colored text, your text box matches the target text box and you are done.

Begin this process by hitting the `Begin transcription' button. This will cause a new `insertText' command to appear in the command log on the right.

You are now in transcription mode. Whatever you say will appear in the `Transcription output` box.

\item \textbf{Editing}

You can fix mistakes in transcription, add formatting, etc. through adding `editText' commands.

Hold down `ctrl' on your keyboard to issue a new `editText' command.

While holding down `ctrl' you will be in edit mode. In this mode, you can manually use mouse-and-keyboard to change the output. However, you must describe the edit you are making before you make it.

Begin by describing your edit using your voice. Whatever you say now will appear in the editText ASR box, but not in the `Transcription output`.

Because the ASR system is imperfect, the textual description may be faulty. Fix any mistakes in the detected speech in the `Gold ASR' box.

Finally, manually edit the `Transcription output' box to correspond the effect of your edit command.

Note: It is important that you vocalize your change before making any edits to either `Gold ASR' or `Transcription output', as the ASR system stops recording as soon as you click into either one of these boxes.

\item \textbf{Undoing, Reseting, Submitting, \& Saving}
You can click on previous commands in the command log to revisit them. Note that if you edit the output associated with a `editText' prior in the history, you will erase the changes associated with subsequent `editText' operations.

If you would like to undo some portion of command log, you can use the `Delete Selected Command \& Afterwards` button. Simply click on the first command you would like to remove, then click the button to remove that command and all commands after it.

You can clear the entire command log by hitting "Reset".

If you would like to work on transcribing another target, use the green arrow keys below the target. This will present you with a new target while saving progress on your current target. To delete a target prompt, press the red `X'.

Once you are done editing, click "Submit" button.

Please double-check each command before submission! In particular, commands will appear red if they are potentially problematic (e.g. they are not associated with any change to the underlying text). Please check to make sure there are no red commands that you do not intend to be there!
\end{enumerate}

\subsection{Target Text Preprocessing}
\label{sec:app_targets}
For replicating \textbf{Enron} emails, we process emails from the  
Enron Email Dataset to create our target final states.
We break the email threads into individual emails, filtering out email headers and non-well-formed emails (emails that are either less than 50 characters or more than 5000 characters long, or contain too many difficult-to-specify non-English symbols).
Annotators also had the option to skip annotating certain emails, if they found the email too difficult to annotate.

\subsection{Annotation Programs}
\label{sec:app_lispress}
Examples of programs can be found below:
\begin{enumerate}
\item ASR: \textit{Lower case the W in the word when.} \\
Program:
\begin{small}
\begin{verbatim}
(lowercase
  (theText
    (and
      (like "W")
      (in
        (theText
          (and
            (word)
            (like "when")))))))
\end{verbatim}
\end{small}

\item ASR: \textit{Get rid of the space in between the two words off site and replace that with a -.} \\
Program:
\begin{small}
\begin{verbatim}
(replace
  (theText
    (and
      (like " ")
      (between
        (theText (like "off"))
        (theText (like "site")))))
  "-")
\end{verbatim}
\end{small}

\end{enumerate}

\subsection{Dataset Analysis}
\label{sec:app_data_breakdown}
\begin{table}[]
    \centering
    \small
    \begin{tabular}{@{}lll@{}}
    \toprule
        \textbf{Actions} & \multicolumn{2}{l}{\textbf{Constraints \& Combinators}}  \\
    \midrule
    combineSentences & union & between \\
    parenthesize & or & endsWith \\
    allCaps & and  & at \\
    do & in & atStart \\
    respell & nthToLast & atEnd \\
    delete & nth & exactly \\
    spell & findAll & hasSubstring \\
    capitalize & thePosition & passage \\
    combine & theText & line \\
    quote & empty & sentence \\
    lowercase & extra & parenthetical \\
    move & nextTo & phrase \\
    moveCursor & take & word \\
    replace & contains & letter \\
    insert & before & text \\
    correction & after & like \\
    & startsWith & alwaysTrue \\
    \bottomrule
    \end{tabular}
    \caption{List of functions present in~\ourdataset.}
    \label{tab:data_cmd_diversity}
\end{table}
\begin{table}[]
    \centering
    \small
    \begin{tabular}{@{}lrr@{}}
    \toprule
        \textbf{Command action} & \textbf{\# of distinct} &  \textbf{\# of distinct} \\
        & \textbf{first tokens} & \textbf{first tokens} \\ 
        & \textbf{(\ourdataset)} & \textbf{(DNS)} \\
    \midrule
        capitalize & 12 & 2 \\
        replace & 83 & - \\
        delete & 22 & 5* \\
        quote & 2 & 1 \\
        parenthesize & 3 & 1 \\
        do & 44 & - \\
        insert & 51 & 1 \\
        correction & 22 & 1 \\
        lowercase & 12 & 1 \\
        allCaps & 8 & 1 \\
        spell & 17 & 1 \\
        move & 3 & - \\
        respell & 1 & - \\
        combineSentences & 7 & - \\
        moveCursor & 3 & 1 \\
        combine & 1 & - \\
    \bottomrule
    \end{tabular}
    \caption{Number of ways to invoke various commands, in terms of number of distinct first tokens used to invoke that command. Second column shows the number of distinct first invokation tokens as present in~\ourdataset, while third column shows the number of distinct first invokation tokens for comparable commands supported by DNS. \\
    \small{*Counting \textit{undo}, \textit{backspace}, and \textit{scratch that} as delete commands, despite being less general than our delete functionality (can only delete most recent tokens).}}
    \label{tab:data_invoke_diversity}
\end{table}
To assess the diversity of \textit{state changes}, we quantify the number of distinct \textit{actions}, \textit{constraints}, and \textit{constraint combinators} (see~\cref{sec:program}) that appear in the annotated programs.
In~\cref{tab:data_cmd_diversity}, we list out all actions, constraints, and constraint combinators present in~\ourdataset. 
\ourdataset 
contains at least 15 types of actions (and allows for action composition with sequential chaining operation \verb|do|), with 
34 types of constraint and constraint combinators.

In~\cref{tab:data_invoke_diversity}, we approximate the invocation diversity represented in~\ourdataset, by measuring the number of distinct first tokens used to invoke each type of actions. For actions that overlap in function with ones supported by DNS, we also report a similar diversity metric against the full set of trigger words supported by DNS.\footnote{\url{https://www.nuance.com/asset/en_us/collateral/dragon/command-cheat-sheet/ct-dragon-naturally-speaking-en-us.pdf}}

\input{sections/online_description}

\section{Model Training Details}
\label{sec:app_training}
In this section, we describe how we trained each component of the system. See~\cref{sec:model} for a description of the inputs, outputs, and architecture of each model.
Our final system is \textit{incremental}, able to process both partial and final ASR results.

\subsection{Segmentation Model}
\label{sec:app_seg_training}
We use BIOES for the segmentation model.
Note that we cannot just predict a binary command/dictation tag for each token, because it would be unable to discern two consecutive commands from one continuous command. Thus, we need to use
\textbf{B}
to specify the beginning of a new command \op.
\textbf{E} is also necessary for the model to predict whether the final \op, in particular, is an incomplete and ongoing (requiring the ASR repair model to predict the future completion) or complete (requiring the ASR repair model to only correct errors).

We expect in the final online version of the end-to-end system, the segmentation model will:
1. run often, being able to accept and segment both partial and final ASR results, 
2. run on only the most recent ASR, to avoid completely resegmenting an entire document that's been transcribed. 
Thus, we construct the training data for this model in a way to simulate these conditions.
We extract all sequences of turns of length between 1 -- 4 from~\ourdataset~(capping to at most 4 for condition 2), take their \utts $u$, and concatenate them to simulate $\allUttsConcated$, asking the model to segment them back into their individual $u$.
For the final turn of each chosen sequence, we include in the training data both the final ASR result and all partial ASR results. 
We fine-tune on this data with a learning rate of 1e-4 and batch size of 4 until convergence.

\subsection{ASR Repair \& Interpretation Models}
\label{sec:app_interpret_training}
Below we describe the concrete implementations and training details of each model:
\paragraph{T5}
In the T5 implementation, both
$\repairModel$
and
$\mathcal{M}_\texttt{INT}$ are
T5-base encoder-decoder models.

As described in~\cref{sec:dataset_details}, we do not have annotations of programs for the full training split. Thus, we automatically generate the missing programs using GPT3.

We have an initial training reservoir that consists solely of data points with program annotations $\mathcal{D}_\text{annot}$.
For each example in the remaining training set,
we retrieve a subset of samples from $\mathcal{D}_\text{annot}$ to form the prompt.
We also use GPT3 for this retrieval step\footnote{we compute similarity between two prompts by looking at the the similarity over next-token distributions when conditioned on each of the prompts}.

We then annotate programs in the remaining training set in an iterative manner: as new programs are annotated, we use the execution engine to check whether it executes to the correct end state, and if so, we add it to $\mathcal{D}_\text{annot}$, such that future examples can include these programs in their prompt.

\paragraph{GPT3}
In the GPT3 implementation, both the ASR repair and interpretation steps occur in a single inference step,
with GPT3 being prompted to predict both outputs in sequence.
Specifically, it is prompted with:

\begin{small}
\begin{quote}
\texttt{[Input State:]} \\
$d_{i-1}$ \\
\texttt{[Utterance ASR:]} $u'_i$ \\
\texttt{[Gold Utterance:]} \colorbox{yellow}{$u_i$} \\
\colorbox{yellow}{\texttt{[Final State:]}} \\
\colorbox{yellow}{$d_i$}
\end{quote}
\end{small}

The model is shown demonstrations in this format from the training data, then asked to infer, for each test sample, the highlighted portions from the non-highlighted portions.

In the setting that we are predicting programs instead of end states, the final 2 lines are replaced with

\begin{small}
\begin{quote}
\colorbox{yellow}{\texttt{[Lispress:]} $\ell_i$}
\end{quote}
\end{small}

\section{Results}
\label{sec:app_results}
\begin{table}[]
    \centering
    \small
    \begin{tabular}{@{}llrr@{}}
    \toprule
         & \textbf{Metric} & \textbf{T5} & \textbf{GPT3} \\
        \midrule
        \multirow{1}{*}{ASR Repair} & EM & $47.3$ & $70.7$ \\
        \midrule
        \multirow{3}{*}{Interpretation} & Program EM & $36.1$ & - \\ 
        & State EM & $33.7$ & $54.2$ \\ 
        \bottomrule
    \end{tabular}
    \caption{We evaluate the ASR repair and interpretation components in isolation.
    We experiment with a fine-tuned T5 vs. a prompted GPT3 model.}
    \label{tab:app_results}
\end{table}

\subsection{Segmentation}\label{sec:app_results_segmentation}
We run all the error analyses in this section on a model trained and tested exclusively on the Replicate doc task (where annotators were asked to replicate emails from the Enron Email Dataset).

We do not evaluate the segmentation model on all of the transcripts that arise during a trajectory, many of which are prefixes of one another.  Doing so would pay too little attention to the later segments of the trajectory.  (F1 measure on the final transcript will weight all of the segments equally, but F1 measure on the earlier transcripts does not consider the later segments at all.) 

Instead, we create an evaluation set of shorter transcripts.  For each trajectory, we form its final full transcript by concatenating all of its final \asr results.  Each sequence of up to 4 consecutive gold segments of this full transcript is concatenated to form a short transcript that the segmentation model should split back into its gold segments.  For example, if the full transcript consists of 8 gold segments, then it will have $8+7+6+5$ evaluation examples of 1 to 4 segments each.

\subsubsection{Error Analysis}
\label{sec:app_seg_errors}
Below, we list some examples of segmentation errors ([$\cdot$] is used to specify \utt boundaries, \hl{yellow-highlighted} segments correspond to command \ops, while non-highlighted segments correspond to dictation \ops).
\begin{enumerate}
\item \textbf{Input:} \textit{Take out the word it. Before the word should. And then replace third with three.} \\
\textbf{True Segmentation:} [\hl{\textit{Take out the word it. Before the word should. And then replace third with three.}}] \\
\textbf{Pred Segmentation:} [\hl{\textit{Take out the word it.}}] [\hl{\textit{Before the word should. And then replace third with three.}}]
\item \textbf{Input:} \textit{You learned. You lie not you learned.} \\
\textbf{True Segmentation:} [\textit{You learned.}] [\hl{\textit{You lie not you learned.}}] \\
\textbf{Pred Segmentation:} [\textit{You learned. You lie not you learned.}]
\item \textbf{Input:} \textit{Skillings calendar is amazingly full! Let's shoot for one of the following.Skillings should be apostrophe s Let's schedule it ASAP.} \\
\textbf{True Segmentation:} [\textit{Skillings calendar is amazingly full! Let's shoot for one of the following.}] [\hl{\textit{Skillings should be apostrophe s}}] [\textit{Let's schedule it ASAP.}] \\
\textbf{Pred Segmentation:} [\textit{Skillings calendar is amazingly full! Let's shoot for one of the following.Skillings should be apostrophe s Let's schedule it ASAP.}]
\end{enumerate}
These examples illustrate two prototypical modes of errors: (i) the ASR system making erroneous judgments about sentence boundary locations, leading the segmentation model astray, and (ii) commands being phrased in ways that disguise them as dictations.
The first example illustrate error type (i): the ASR system oversegments the input (which should've been a single sentence) into three separate sentences, consequently leading the segmentation system to believe ``\textit{Take out the word it}'' and ``\textit{Before the word should...}'' are separate commands.
The second example illustrates error type (ii): ``\textit{You lie not you learned.}'' is supposed to be a replace command indicating ``\textit{You learned}'' should be replaced with ``\textit{You lie}'', but it is not phrased as an explicitly command.
Finally, the third example illustrates both error types: we see that the ASR system undersegments the input and combines the sentence ``\textit{Skillings should be apostrophe s}'' with the sentence ``\textit{Let's schedule it ASAP}'' without a period. Combined with the fact that ``\textit{Skillings should be apostrophe s}'' is not issued explicitly as a command, this confuses the segmentation model into thinking that it is in fact part of the dictation.

\subsubsection{Success Cases: Fixing Erroneous Segmentations}
\label{sec:app_seg_successes}
The above examples illustrated certain cases where the segmentation model was misled by erroneous ASR judgments about sentence boundary locations. 
In some cases, however, the segmentation model is able to fix these judgements, as shown below:
\begin{enumerate}
    \item \textbf{Input:} \textit{Take out the extra space. In between the two words, but and should.} \\
    \textbf{True/pred Segmentation:} [\hl{\textit{Take out the extra space. In between the two words, but and should.}}]
    \item \textbf{Input:} \textit{Replace the period. With a comma after restructuring.} \\
    \textbf{True/pred Segmentation:} [\hl{\textit{Replace the period. With a comma after restructuring.}}]
\end{enumerate}

\subsection{ASR Repair}
\label{sec:app_eval_asr}
\paragraph{Metrics}
To measure the ASR repair step in isolation, we take noisy utterances $u_i$ corresponding to each command and measure to what extent we are able to reconstruct the ground-truth utterance. We measure the percent of $u_i$ for which our predicted repaired utterance exactly matches the ground-truth utterance (\textbf{EM}).

\paragraph{Results} From~\cref{tab:app_results}, we see that the GPT3 model is much better at repairing speech disfluencies and ASR errors than the T5 model, achieving 70\% EM. We suspect this is due to the fact that GPT3 was pretrained on much more (English) language data than T5, 
giving GPT3 a greater ability to produce grammatically coherent and permissible English sentences,
and likely also 
a better sense of
common disfluencies. 

\paragraph{Qualitative Analysis}
Recall that we designed the ASR repair step to condition on not just the utterance $u_i$ but the state $d_{i-1}$. This allows it take $d_{i-1}$ into account when repairing $u_i$. 

For example, when given the following utterance:
\begin{quote}
\textit{Delete the period after events.}
\end{quote}
An ASR repair model that looks at ASR alone may not see any issue with this utterance.
However, given the document state:
\begin{quote}
Eric, I shall be glad to talk to you about it. The first three days of the next week would work for me. Vince.
\end{quote}
(note the word \textit{events} does not appear anywhere in this text),
the ASR repair model
realizes that the actual utterance should've been,
\begin{quote}
\textit{Delete the period after Vince.}
\end{quote}
Indeed, the T5 ASR repair model is able to make the appropriate correction to this utterance.

\subsection{Interpretation}
\label{sec:app_eval_interpret}
\paragraph{Metrics}
To measure the interpretation step in isolation, we take normalized utterances $u'_i$ corresponding to each command and measure to how well the interpretation model is able to reconstruct the ground-truth final state for the command $d_i$. 
We use the same set of metrics described in~\cref{sec:eval_repair_interpret} (state EM, program EM). However, insteading of feeding the interpretation model ASR repair results, we feed in ground-truth utterances $u$.

\paragraph{Results}
We evaluate a T5 interpretation model that produces programs (which is then executed by our execution engine) and a GPT3 interpretation model that directly generates states. Results are reported in~\cref{tab:app_results}.

We can also compare these isolated interpretation results with the joint ASR and interpretation results reported in~\cref{tab:results}.
Due to error propagation, the T5 model is $\sim$5--8\% worse 
when asked to jointly perform ASR repair and interpretation from noisy ASR, than when simply asked to interpret normalized utterances.
Surprisingly however, the GPT3 model performs nearly as well in the joint evaluation as the isolated evaluation.
We suspect that even if the GPT3 ASR repair model does return the exactly normalized utterances, it is still able to reconstruct a semantically equivalent/similar command. 

\section{Infrastructure and Reproducibility}
We trained 220M-parameter T5-base model on a single NVIDIA Tesla A100 GPU machine.
Each training run for each component of the model took at most a few hours ($<$8).
We also prompted a 12B-parameter GPT3 model.

We used PyTorch~\cite{pytorch} and Huggingface Transformers~\cite{wolf-etal-2020-transformers} for implementing and training T5-base models. We use OpenAI's API\footnote{\url{https://beta.openai.com/}} for querying GPT3. We use the \texttt{text-davinci-003} model.

%% file: sections/online_description.tex
\section{Running Online}\label{sec:commitpoint}
\label{sec:app_online}
When running the system online in real time, we
must consider efficiency and usability.
We introduce a ``commit point'' that signifies that the system cannot re-segment, re-normalize, or re-interpret anything before that point.
We only want to consider recent \asrs because 
the system quickly becomes inefficient as the dialogue length grows
(the interpretation step, which is the bottleneck of the system, must run for every single command.)
Furthermore, users often refer to and correct only recent dictations and commands; reverting early changes can have potentially large and undesirable downstream effects, leaving users potentially highly confused and frustrated.

Concretely, the commit point is implemented as the system treating the document state at that point as the new ``initial state,'' so that it is unable to access segments and the history of document states from before that point. We implement this point so that it must coincide with the end of a final \asr.
We feed into the system this state as the initial state, and the entire sequence of \asrs
starting from that point.
All dictations and command \ops returned by the model are executed in sequence from the commit point.

We decide to set a commit point based on 
system confidence and time since last commit. 
System confidence is derived from the confidences of each 
component model at each step of the prediction.
We measure the system confidence of the \textit{end state} predicted by the system, by summing the log-probabilities of: 1. the segmentation model result, (summing the log-probabilities of each BIOES tag predicted for each token), 
2. the ASR repair model result for each command (log-probability of the resulting sentence), 3. the interpretation model result for each command (the log-probability of the end state or program).
Once the system confidence exceeds a threshold $\tau_\text{commit}$, we decide to commit immediately at that point.
Otherwise, if we have obtained
more than 4 final \asrs
since the last commit, we must commit at our most confident point from within the last 4 turns.